# Dense Residual Network: Enhancing Global Dense Feature Flow for Character Recognition

Zhao Zhang[1,2], Zemin Tang[3], Yang Wang[1,2], Zheng Zhang[4], Choujun Zhan[5], Zhengjun Zha[6], Meng Wang[1,2]

[1] School of Computer Science and Information Engineering, Hefei University of Technology, Hefei 230009, China
[2] Key Laboratory of Knowledge Engineering with Big Data (Ministry of Education) & Intelligent Interconnected Systems Laboratory of Anhui Province, Hefei University of Technology, Hefei 230009, China
[3] School of Computer Science and Technology, Soochow University, Suzhou 215006, China
[4] Bio-Computing Research Center, Harbin Institute of Technology (Shenzhen), Shenzhen, China
[5] School of Computer, South China Normal University, Guangzhou 510631, China
[6] Deparmtment of Computer Science and Technology, University of Science and Technology of China, Hefei, China

***Abstract*— Deep Convolutional Neural Networks (CNNs), such as Dense Convolutional Network (DenseNet), have achieved great success for image representation learning by capturing deep hierarchical features. However, most existing network architectures of simply stacking the convolutional layers fail to enable them to fully discover local and global feature information between layers. In this paper, we mainly investigate how to enhance the local and global feature learning abilities of DenseNet by fully exploiting the hierarchical features from all convolutional layers. Technically, we propose an effective convolutional deep model termed Dense Residual Network (DRN) for the task of optical character recognition. To define DRN, we propose a refined residual dense block (r-RDB) to retain the ability of local feature fusion and local residual learning of original RDB, which can reduce the computing efforts of inner layers at the same time. After fully capturing local residual dense features, we utilize the sum operation and several r-RDBs to construct a new block termed global dense block (GDB) by imitating the construction of dense blocks to adaptively learn global dense residual features in a holistic way. Finally, we use two convolutional layers to design a down-sampling block to reduce the global feature size and extract more informative deeper features. Extensive results show that our DRN can deliver enhanced results, compared with other related deep models.**

## I. Introduction

Deep CNNs with multiple layers have made significant progress and achieved great success in many vision tasks, such as image recognition, speech recognition and video person recognition, by learning deeper representations and hierarchical information [47-52]. This success has also been demonstrated by the optical character recognition (OCR) that reads the scene text in images and predicts a sequence of characters from the machine generated texts [12-14][33][36-38]. OCR has been widely applied in various applications, e.g., road sign recognition, identification, license plate recognition and assistive service for the blind.

For the task of OCR, two crucial sub-tasks are text line detection and text recognition [12-14]. The first task is to extract the text regions from images and the second one is to recognize the textual contents of the identified regions. In this paper, we mainly discuss the task of text recognition rather than text line detection. As different images have complicated backgrounds and complex contents, OCR is still a challenging task. To tackle this task, many OCR models have been proposed, e.g., arbitrary orientation network (AON) [2], end-to-end trainable scene text recognition system (ESIR) [1] and the convolutional recurrent neural network (CRNN) [3]. AON presented an arbitrary orientation network to recognize the oriented texts arbitrarily and achieved impressing results on both irregular and regular texts from images. ESIR has designed a novel line-fitting transformation to estimate the pose of text lines in scenes and has developed an iterative rectification framework for the scene text recognition. CRNN is the combination of two prominent neural networks, i.e., CNNs and Recurrent Neural Networks (RNNs) [5-7]. More specifically, CNNs are used to extract information of images [46], RNNs are used to predict the label distribution of each frame and Connectionist Temporal Classification (CTC) module [8] is also used to transform these predictions into the final label sequence. Note that recent work also revealed that even without the recurrent layers, the simplified models can still achieve the promising results with higher efficiency [9][34]. As such, the framework of CNNs plus CTC is a feasible and efficient solution. To extract features in convolutional layers, many existing convolution networks can be used, e.g., Dense Convolutional Network (DenseNet) [4], Residual Network (ResNet) [10] and Residual Dense Network (RDN) [11].

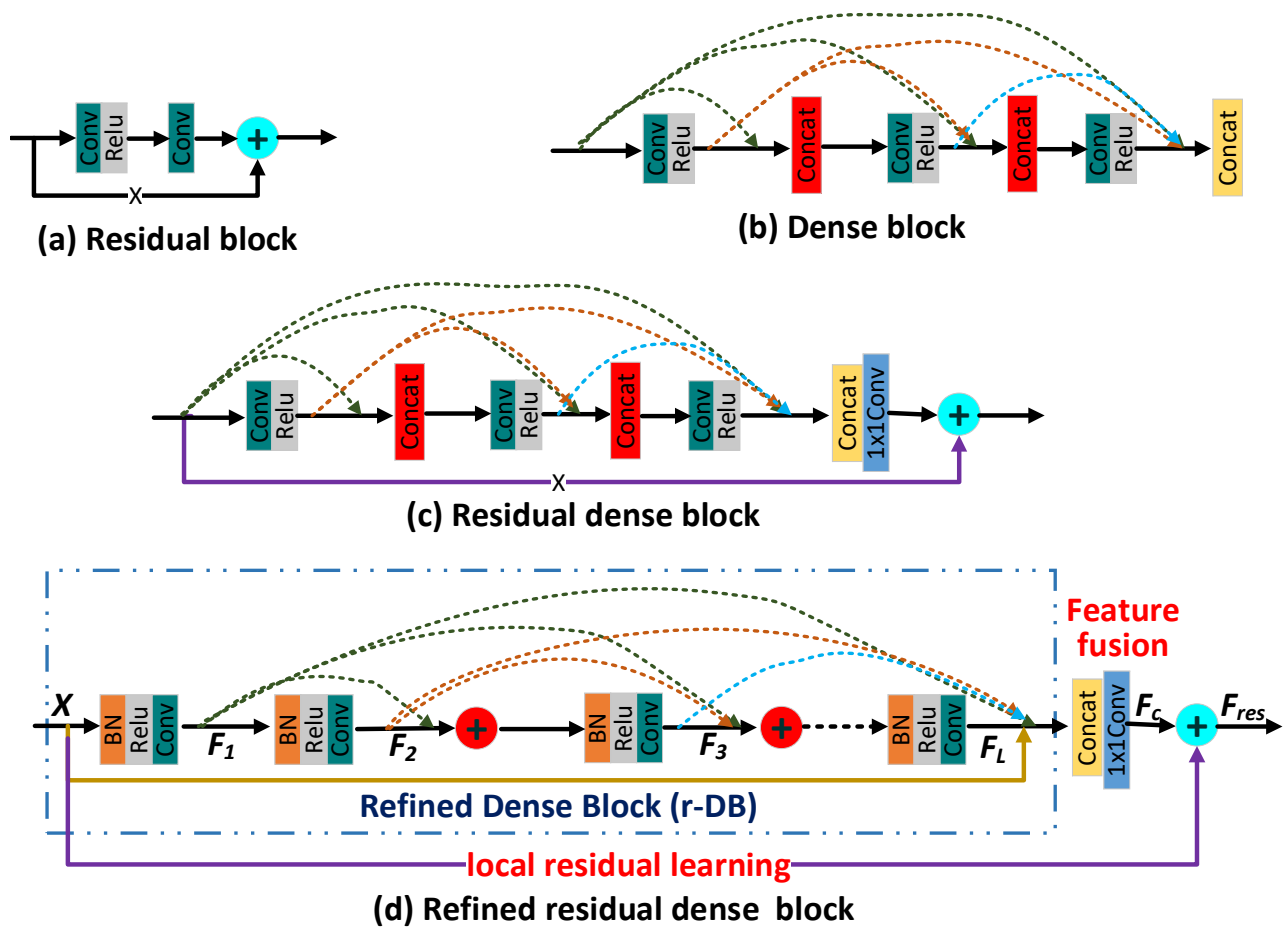

Fig. 1. The structures of (a) residual block, (b) dense block, (c) residual dense block (RDB), and (d) our refined residual dense block (r-RDB).

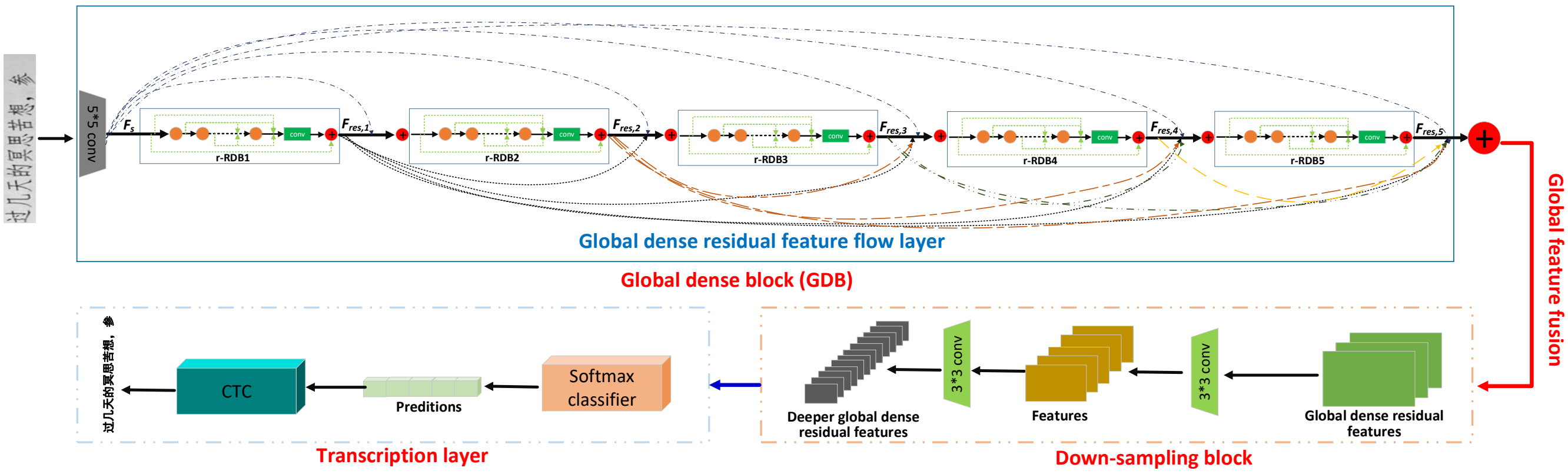

Fig. 2. The learning architecture of our proposed framework for text recognition from images.

It is noteworthy that the texts in images have different scales, angles of view and aspect ratios [11]. Although the hierarchical deep features extracted by a deep network could provide more clues for recognition, most existing CNNs based models usually neglect to use hierarchical features for recognition or only focus on learning local hierarchical features. For example, the dense block of DenseNet connects different inside layers tightly, so it has a strong ability in learning local features due to its intrinsic structures. However, like most existing CNN models, DenseNet also stacks the dense blocks [4][45] and transition blocks simply, which neglects the global properties of features. In addition, the way of combining features by concatenating them in DenseNet will bring about sharp increase in terms of the channel number of input features in each layer and huge computing efforts with dense block getting deeper, which will restrict the depth of the deep networks using the dense blocks. RDN proposed a residual dense block (RDB) [11] for image super-resolution. It should be noted that RDB has some obvious advantages compared with the residual block of ResNet and dense block. Specifically, the structure of RDB contains the dense connected layers, local feature fusion (LFF) and local residual learning (LRL) modules, which can fully capture local hierarchical features and learn the local residual dense features. However, RDB uses the standard dense blocks, so it will have the same disadvantages as the dense blocks in terms of higher computing efforts.

In this paper, we propose a deep convolution network model with CTC, which can fully use and enhance the local and global hierarchical features from the text images and reduce the computing efforts at the same time. In summary, the main contributions of this paper are presented as follows:

(1) Technically, we propose a new effective deep representation learning and character recognition network, i.e., dense residual network (DRN). The proposed DRN can fully use all the local and global hierarchical features, and moreover enhance the global dense feature flow, while global features are usually ignored in existing models. That is, DRN can enhance the local and global feature learning by exploiting hierarchical features from all convolutional layers fully.

(2) We also propose a RDB based refined residual dense block (r-RDB), which can clearly retain the ability of local feature fusion and local residual learning of the original RDB, but also reduce the computing efforts of inner layers by refining the structures of the representation block at the same time.

(3) After learning the multi-level local residual dense features by r-RDB, we use the sum operation and several r-RDBs to construct a new global dense block (GDB). GDB is clearly designed by imitating the dense block, but it can adaptively learn global dense residual features by connecting features of all the r-RDBs tightly in the form of dense connection in a holistic way. Note that discovering global deep features is the major innovation of this paper, but has been ignored by the vast majority of existing deep networks.

(4) We also design a down-sampling block with two convolutional layers of stride 2 to reduce the size of global features and extract more informative deeper global features due to having more kernel channels in this block. This block can also avoid important feature information loss and make the parameters of the whole framework learnable. To further reduce the computing effort and improve the efficiency, we use the depth-wise separable convolution to replace the traditional convolution.

The paper is outlined as follows. Section II briefly reviews the related models. In Section III, we introduce the r-RDB and global dense block (GDB). We present the deep framework of our fast dense residual network (DRN) in In Section IV. Section V shows the experimental setting and results. In Section VI, we describe the conclusion and future work.

## II. Related Work

In this section, we briefly introduce the residual block, dense block and residual dense block, which are closely related to our proposed global dense block and deep network model.

**Residual block in ResNet.** ResNet mainly solves the problem of network degradation [10]. Fig.1(a) shows the structure of one residual block. Let $x$, $F(x)$ and $H(x)$ denote the input features, output features without and with short connections respectively, we can easily obtain the following formulas:

$$\begin{aligned} F(x) &= w_2(\mathrm{ReLU}(w_1 x)) \\ H(x) &= F(x)+x \end{aligned}, \qquad (1)$$

where $w_1$ and $w_2$ are the weights for the convolutional layers, $\mathrm{ReLU}(w_1 x)$ is the function of Rectified Liner Units (ReLU) [18]. The identical maps can be constructed in two cases that $F(x) = x$ without short connections or $F(x) = H(x) - x = 0$ with short connections. The optimization with a short connection is easier in the process of training. In addition, the short connection in the residual block can also help enhance the feature flow, which improves the ability of fusing local features to a certain extent. However, from the aspect of global feature learning, the representation ability of ResNet will be limited.

**Dense block in DenseNet.** DenseNet [4] mainly creates short paths between layers, which can alleviate the issue of gradient dis-appearance, strengthen feature flow and encourage feature reuse. In this process, a simple connectivity pattern called dense block is derived. To ensure maximum information flow between layers, all the layers with matching feature-map sizes in a dense block are connected directly with each other, and the related features are combined by concatenating them as follows:

$$X_i = \phi\left(\left[X_0, X_1, ..., X_{i-1}\right]\right), \tag{2}$$

where $X_i$ are features of the *i-th* layer, $X_0$ are input features and $\phi(\bullet)$ is the according function of convolutional layer. To preserve the feed-forward nature, each layer obtains additional inputs from preceding layers and passes on its own feature-maps to the subsequent layers. Fig.1 (b) shows the layout of one dense block. Note that DenseNet uses many short connections to enhance the feature flow, thus it will have a strong feature learningability. As far as the local feature fusion is concerned, the designed dense block is innovative and the module has a strong ability of local feature fusion and learning. However, like the ResNet, DenseNet also fail to learn global features, since it also just stacks the dense blocks one by one. Besides, with the increase of the number of convolutional layers in dense block, the computing efforts will increase dramatically, which will directly limit the depth of neural network.

**Residual dense block (RDB) in RDN.** The residual dense network (RDN) [11] aims to make full use of all the hierarchical features from images. The core idea of RDB is to utilize both residual learning and dense block to improve the feature learning ability by jointly enhancing the information flow, learning residual features and improving the local feature fusion. As such, RDB can fully learn local hierarchical features and obtain informative residual dense features. Fig.1(c) illustrates the structure of one RDB. It should be noted that RDB also utilizes the concatenating operation similarly as the dense block to combine the features of former RDB and current layers, and use the residual outside dense block to enhance the representation learning. As such, RDB will also suffer from huge computing cost as the dense block. In addition to making full use of local features with RDB, RDN has also made its own work in global feature fusion learning. However, by comparing with the structures of the dense connection in RDB, RDN just simply combine all the features from different RDBs in the form of concatenating for the global feature fusion. As such, the global feature learning ability of the RDN is still not enough [11], where global information flow between layers are not fully discovered.

## III. Dense Residual Network (DRN) for Character Recognition

We first describe the network architecture of our DRN in Fig.2, where it consists of three major parts, i.e., a global dense block (GDB), a down-sampling block and a transcription layer. It is noted that traditional CNN models usually neglect to use hierarchical features for image representation or only focus on local hierarchical features. It should be noted that all the convolution operations in our DRN refer to an operation group including the batch normalization (BN) [17], rectified liner units (ReLU) [18] and the depth-wise separable convolution [19]. Note that the major reason why we put GDB before the downsampling module is to make use of more abundant feature information as much as possible. As we know, downsampling will reduce the size of features. Although more channels can be set to increase the amount of information after convolution, a considerable amount of information will be lost for each channel feature. Next, we briefly introduce the global dense block (GDB), down-sampling block and transcription layer in our DRN.

### *A. Global Dense Block (GDB)*

To enhance the global feature learning on the premise of fully learning local features, we define a fast RDB and a global dense block, which will be detailed in the next section.

### *B. Down-sampling Block*

We first use GDB to compute the global dense residual features. After that, we use two convolution layers with stride 2 to reduce the global feature size. We also set more channels for the convolutional kernels in the down-sampling block to learn more informative global dense residual features. As such, we can find that the feature channels turn out more and more with gradually decreasing feature size in the down-sampling block in Fig.2.

### *C. Transcription layer*

This layer is used to transform the prediction of each frame into the final label sequence, which includes the soft-max and CTC. The soft-max function [20] is utilized to output the predictions of the down-sampling block. CTC plays the role in transforming these predictions into the final label sequence. In our network, CTC needs to input data of each column of a picture containing text as a sequence and outputs the corresponding characters.

## IV. Refined Residual Dense Block (r-RDB) and Global Dense Block (GDB)

We mainly introduce the refined residual dense block (r-RDB) and global dense block (GDB). In what follows, we show their definitions and also illustrate their structures.

### *A. Refined residual dense block (r-RDB)*

We first describe the structure of r-RDB that is designed based on the RDBs. RDB includes a dense block, a 1*1 convolution layer and a sum operation for the residual learning. To design the structure of r-RDB, we clearly borrow the idea of reducing the computing efforts and weight size of the dense blocks by changing the way of combining inner features [16] into RDB, so that the computational cost and weight size can be greatly reduced while retaining the advantages of RDB in local feature fusion and residual learning. Specifically, r-RDB is also constructed by redefining and designing the way of combining internal features of inner dense block as the refined dense block (r-DB) [16]. Generally, since the channel number of input features is more than that of the inner layers in dense blocks, we similarly reduce the computing cost by applying the sum operation to combine features instead of concatenating for all the inner layers in the dense blocks, except for the input and output layer, which is performed as follows:

$$F_1=\phi\left(X\right) \\ F_i=\phi\left(F_1+...+F_{i-1}\right),\ i>1 , \tag{3}$$

where $F_i$ represents features of the $i$-th layer, $X$ represents the input features and $\phi(\bullet)$ denotes the function used in the convolutional layer. Since we construct output features by concatenating all features of different layers, we can fully utilize feature information of all layers and can also ensure that our feature maps have the same size and same number of channels as those of the original dense block in RDB. As such, we can find that the channel number of input features grows in the original dense blocks but keeps unchanged in r-RDB for those middle layers. According to [16], the lightweight version can clearly reduce the computing cost and weight size of each dense block to ($1/L$, $2/L$), where $L$ is the number of layers in dense block. Since the computation in RDB mainly depends on the inner dense blocks, while the 1*1 convolution kernels and sum operation for the residual learning only takes a small share of the computation, so r-RDB can reduce the computation cost of the inner layers of RDB, similarly as r-DB. Moreover, r-RDB can also preserve the ability in residual dense feature learning of RDB effectively. Fig.1(d) shows the structure of a r-RDB, where it clearly contains fast dense block, local feature fusion and local residual learning, shown below:

**Local feature fusion.** Similar to RDB, we use a 1×1 convolution layer to adaptively control the output information, which can not only extract convolutional features from concatenated features, but also reduce the channel number of fused features for residual learning. The local fusion features are defined as

$$F_c=\varphi\left(\left[X,F_1,F_2,F_3...F_L\right]\right), \tag{4}$$

which is obtained by implementing a 1*1 convolutional operation on the concatenated features from r-DB, $L$ is the number of layers in r-DB, $\varphi(\bullet)$ is the according function of 1*1 convolutional operation, and $[\bullet]$ is the concatenating operation.

**Local residual learning.** To add the input features and fusion features, the sum operation is introduced for residual learning in our r-RDB similarly as RDB. This operation can help us further improve the information flow, improve the network representation ability and finally enhance performance. Since our r-RDB also does the residual learning on dense features, we call these output features as local residual dense features as well, denoted as $F_{res}$ that is defined as follows:

$$F_{res}=X+F_c . \tag{5}$$

## *B. Global Dense Block (GDB)*

We then describe the structure of GDB that is our core module, which is mainly designed to enhance the global feature learning on the premise of fully-learned local features. The structure of GDB is designed by imitating the dense block, but it can jointly extract the shallow features, ensure fast global dense residual feature flow and enable the global feature fusion. To be specific, GDB consists of three parts, including a shallow feature extraction layer that can extract the shallow features, a global dense residual feature flow layer that ensures the fast global dense residual feature flow and a global feature fusion layer that enables the global feature fusion.

**Shallow feature extraction layer.** This layer aims to extract and learn shallow features from the original images containing texts. It also plays the role of down-sampling when encountering with large-size input features. The 5*5 conv in this layer of Fig.2 is a convolutional operation with kernel size being 5*5. The extracted shallow features $F_s$ from input are defined as

$$F_s=H_{dwconv}\left(H_{RELU}\left(H_{BN}\left(X\right)\right)\right), \tag{6}$$

where $H_{dwconv}(\bullet)$ is depth-wise separable convolution operation, $H_{BN}(\bullet)$ is BN operation and $H_{RELU}(\bullet)$ is ReLU operation.

**Global dense residual feature flow layer.** As claimed above, r-RDB can fully obtain the local residual dense features, then we can extract the global dense residual features by this GDB based layer. Like the dense block, all inner layers are connected directly with each other in GDB to ensure maximum information flow between layers. Besides, each layer in GDB obtains additional inputs from all the preceding layers and passes on its own features to all subsequent layers. To ensure the computing efficiency of network, we take the sum operation to combine features in GDB instead of concatenating operation.

**Global feature fusion layer.** Finally, we combine features of all layers of GDB by a sum operation. In this way, we can fully use shallow features and local residual dense features to obtain the final global features. Since r-RDB can extract local residual dense features that are placed in the way of dense connection, we can call the output features of GDB as *global dense residual features*. Let $F_{res,\,i}$ ($i$=1,…,5) and $F_{global}$ denote output features of each r-RDB and GDB, where $F_{res}$ have been shown in formula 5. Here, we present the formula of $F_{global}$ as follows:

$$F_{global}=F_s+\sum_{i=1}^{5}F_{res,\,i} . \tag{7}$$

## *C. Discussion*

In this section, we discuss the relations and differences between the feature leaning abilities of RDN and our DRN.

**Local feature learning.** RDN designs the RDB to extract and learn local residual dense features, while r-RDB also introduces the idea of fast computation into RDB to further improve the efficiency and retain the representation ability of the local dense feature fusion and local residual learning.

**Global feature learning.** How to fully discover and enhance the global feature learning ability is the major problem studied in this work and is also the biggest innovation of our DRN. As described above, most existing CNN models such as ResNet and DenseNet improved the ability of fusing local features, but they stacked their residual blocks and dense blocks one by one to construct their global structures respectively. These will clearly result in the loss of the hierarchical features of each block in theoverall structures, which are in fact important clues for the image representation and recognition. Besides, RDN learns its global fusion features by concatenating features of all the RDBs to improve the performance. However, comparing to the structures of RDB that connects local hierarchical features of all the layers tightly, the overall architecture of RDN does not do well in integrating the global features. Different from other existing models, our DRN not only connects features of all the r-RDBs tightly in the form of dense connection imitating the design of the dense block, but also utilizes the sum operation to reduce the computing efforts. Further, we use two convolution layers with more channels to reduce the feature size and learn deeper global dense residual features. By this way, we can enhance the ability of learning global features for representation learning.

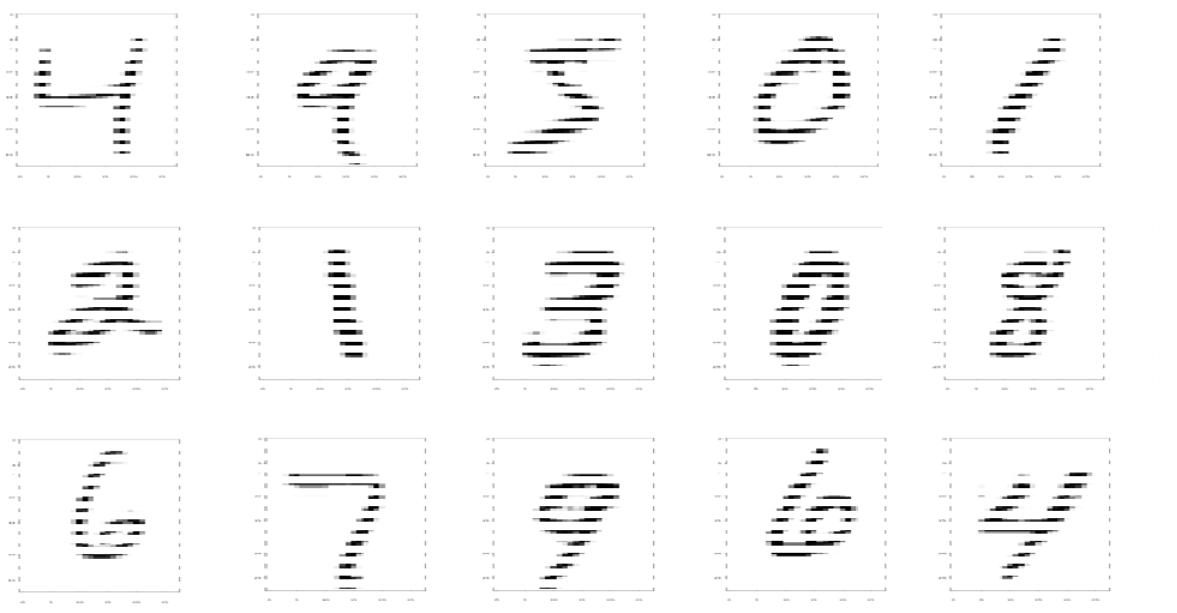

Fig. 3. Illustration of some image examples in MNIST.

TABLE I
COMPARISON RESULTS OF EVALUATED DEEP MODELS ON THE MNIST HANDWRITTEN DIGIT DATABASE.

| Evaluated Frameworks | Accuracy (%) |
|---|---|
| Deep L2-SVM | 99.13% |
| Maxout Network | 99.06% |
| BinaryConnect | 98.71% |
| PCANet-1 | 99.38% |
| gcForest | 99.26% |
| Simple CNN with BaikalCMA loss | 99.47% |
| **DRN (ours)** | **99.67%** |

## V. EXPERIMENTAL RESULTS AND ANALYSIS

We evaluate the performance of our DRN for text image representation and recognition. In this study, we mainly consider two recognition tasks: (1) recognizing the character strings in images; (2) recognizing the handwritten characters from images. For the first task, we compare the results of our DRN with those of several related deep network models, where the CPUs and GPUs of all the evaluated methods in the experiments are Xeon E3 1230 and 1080 Ti respectively and the used convolution architectures are based on the framework of Caffe [21]. A large-scale synthesis Chinese String dataset [22] is used for this evaluation. For the second task, we compare the recognition results with some popular methods on MNIST [23] and HASY [43] for character recognition. It is noteworthy to point out that CTC is required in the first task of character recognition, but is not needed in the second handwritten digits recognition task.

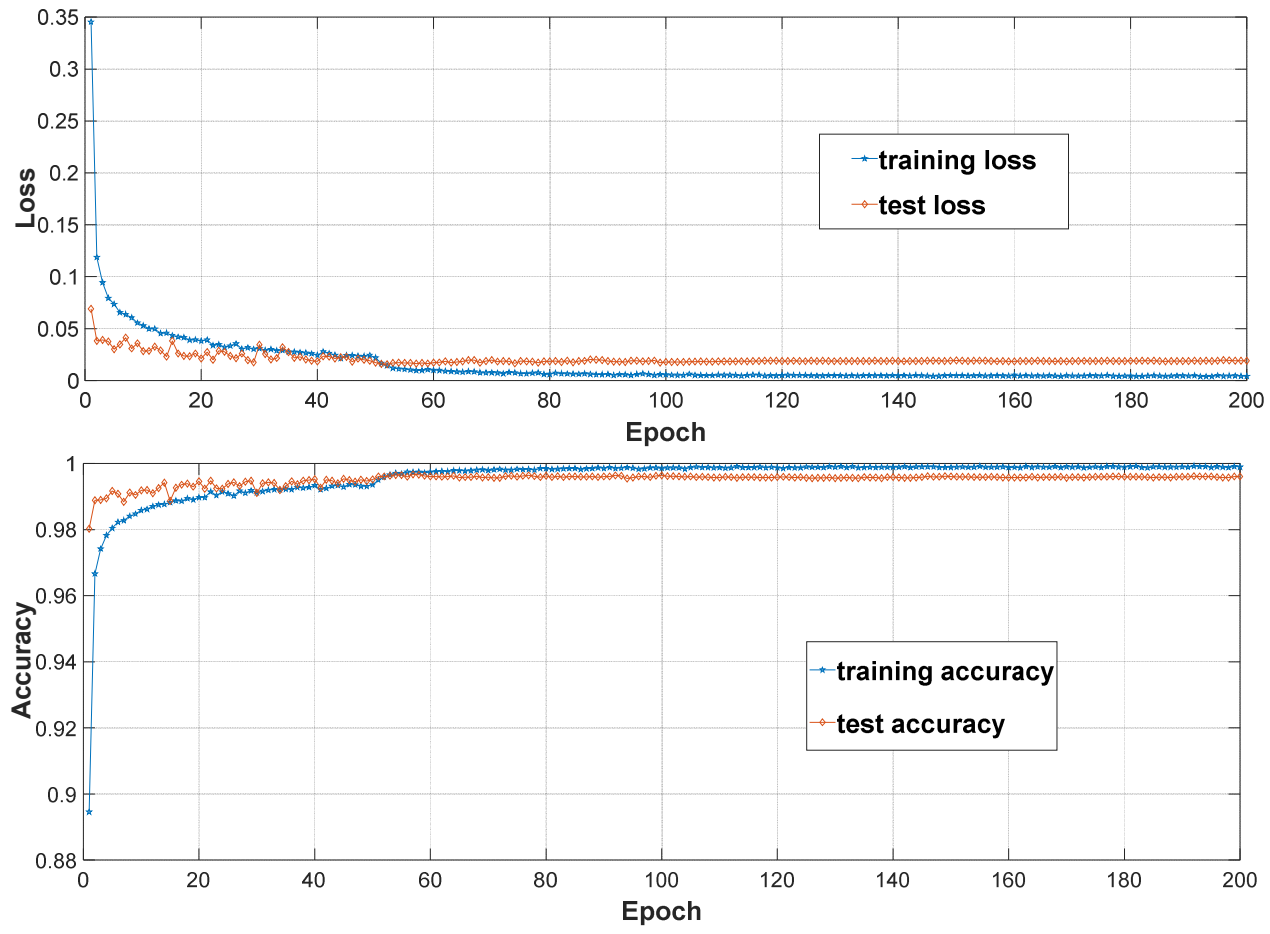

Fig. 4. Training curves of our DRN on MNIST.

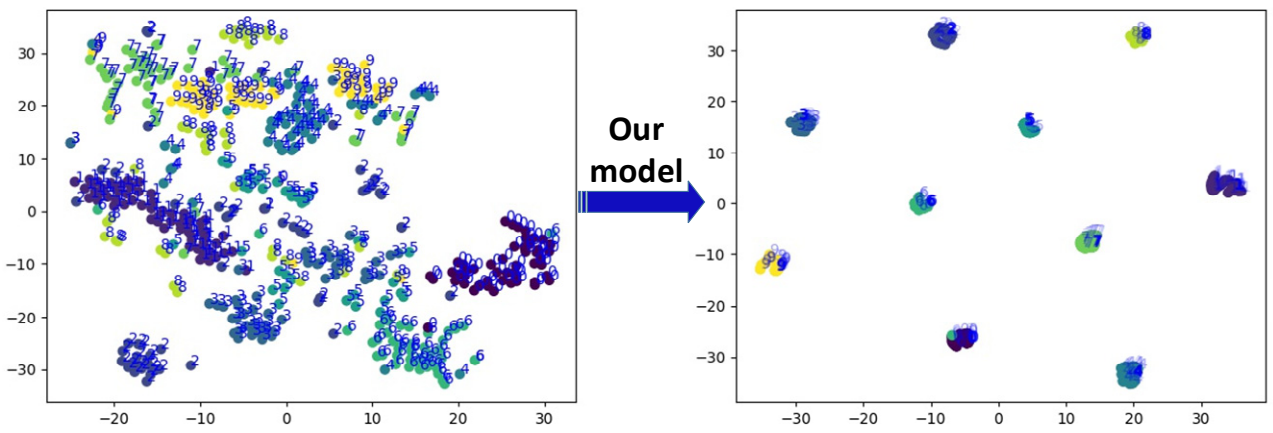

Fig. 5. Distribution of the original MNIST dataset (left) and deep features of MNIST by our DRN (right).

Since the second task only needs to perform the single character recognition, we utilize the convolution parts of DRN, i.e., GDB and the down-sampling block is used to extract the deep dense residual features and then use the soft-max function as a classifier to predict the labels of data samples.

### A. *Handwritten Character Recognition*

**(1) Experiments on MNIST**

We first evaluate DRN for recognizing the handwritten digits using the popular MNIST database. MNIST is a widely-used handwritten digit dataset, where the goal is to classify the images with 28×28 pixels as one of the 10 digital classes. MNIST dataset has 60,000 training samples and 10,000 testing samples. The results on MNIST can evaluate the feature extraction and representation learning ability of a learning model. In Fig.3, we show some image examples of the MNIST dataset.

**Implementation details.** For MNIST, the batch size is set to 128 and the epoch size is 200 in this study. The initial learning rate is set to 0.001, which will be adjusted to 0.0001 at interval between 50 and 100, and to 0.00001 after 100 epochs. Note that we add a fully-connected layer after down-sampling block in our DRN so that the output features can be transformed into the required form of soft-max. To prevent overfitting, we add three dropout layers after 5*5 conv, down-sampling block, and the fully connected layer, and set the value to 0.5, 0.5 and 0.7 in DRN. The performance of our DRN is compared with those of six popular deep models, such as Deep L2-SVM [24], Max-out Network [25], BinaryConnect [26], PCANet-1 [27], gcForest [28] and Simple CNN with BaikalCMA loss.

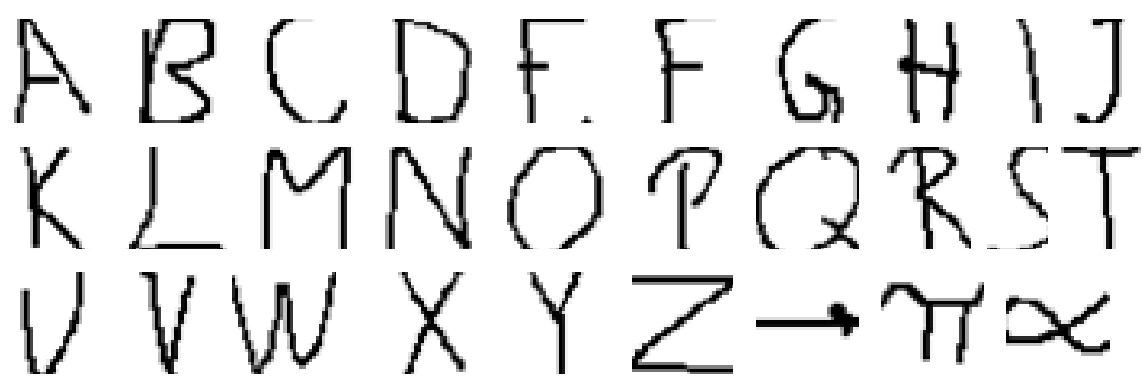

Fig. 6. Illustration of some image examples in HASY.

TABLE II
COMPARISON RESULTS OF EVALUATED DEEP MODELS ON HASY.

| Evaluated Frameworks | Accuracy (%) |
|---|---|
| Random Forest | 62.4% |
| MLP | 62.2% |
| LDA | 46.8% |
| CNN-3 | 78.4% |
| CNN-4 | 80.5% |
| CNN-4A | 81.0% |
| CNN-3+ displacement features | 78.8% |
| CNN-4+ displacement features | 81.4% |
| CNN-4A+ displacement features | 82.3% |
| **DRN (ours)** | **85.0**% |

**Recognition results.** The handwritten recognition results in terms of accuracy on MNIST are described in Table I. We see that DRN can obtain the enhanced results and the recognition accuracy reaches 99.67 and is about 0.2%-0.54% improvement compared with baseline, which implies that the proposed r-RDB and DRN have strong representation and recognition abilities.

**Visualization of training process.** We also show the training curves of our DRN that is trained using the cross-entropy loss. The obtained training curves are shown in Fig.4, where the top figure is the curve of the cross-entropy loss and the bottom one shows the recognition accuracy. We see that the cross-entropy loss is well fitted, which implies that DRN has a strong ability for representation learning and handwritten recognition.

**Visualization of the data distribution.** To further observe the performance of our model, we use t-distributed stochastic neighbor embedding (t-sne) [39] to reduce the dimension of the original MNIST and its features from the last layer of our DRN, so that we can see their distributions. We see from Fig.5 that the learned features by our DRN have enhanced inter-class separation and intra-class compactness, which also proves that our network has a strong representation learning ability.

**(2) Experiments on HASY**

We then evaluate each deep model for recognizing the handwritings of HASY [43]. HASY is a public dataset of single symbols but more challenging, because the number of classes in HASY is more than that in MNIST and there are many similar classes in HASY, i.e., it has 168,233 instances with 369 classes. HASY includes two challenges: a 10-fold cross-validation task for classification and a verification task. This paper focuses on the classification task. Some examples are shown in Fig.6.

**Implementation details.** In the task of handwritten classification with 10-fold cross-validation, the data of HASY is divided into training set and test set in the proportion of 9:1, and 10 datasets which are divided into different images in the same proportion are provided. In the experiment of each dataset, the batch size is set to 64, and the epoch size is set to 20. The initial learning rate is set to 0.001 and will be adjusted to 0.0001 between 10 and 20. A fully-connected layer is added after the down-sampling block in DRN. To prevent the over-fitting, a dropout layer [44] is added after the fully connected layer, and the parameter value is set to 0.5. Average the convergence results of each dataset to obtain the final classification accuracy.

**Recognition results.** We compare our model with five popular deep models, i.e., random forest [40], multi-layer perceptron (MLP) [41], linear discriminant analysis (LDA) [42], CNN-3/4/4A [43] and (CNN-3/4/4A) + displacement features [44]. The random forest uses a large number of decision trees in the integrated classifier; LDA is a linear classifier; MLP is a fully- connected forward neural network; CNN-3/4/4A is a multi-layer CNN model; (CNN-3/4/4A)+ displacement features is a multi-layer CNN model by using displacement features. It can be seen from TABLE II that our proposed DRN outperforms other competitors by delivering better recognition results and the improvement is about 2.7%-22.6% by comparing with other compared methods, which once again implies that the proposed model have stronger representation and recognition abilities.

### *B. Character String Recognition in Images*

In this section, we evaluate each deep framework for recognizing the texts in images using the synthetic Chinese string dataset. This dataset is generated from Chinese corpus, including news and classical Chinese and by changing fonts, sizes, gray levels, blurring, perspective and stretching [22]. The dictionary has about 5990 characters, including Chinese, punctuation, English and numbers. Each sample is fixed to 10 characters, and characters are randomly intercepted from the corpus. The resolution of the pictures is unified to 280×32. A total of about 3 million 600 thousand images are generated, which are divided into a training set and a test set according to 9:1. Fig.7 shows some image examples of the Chinese string dataset.

**Implementation details.** We use the stochastic gradient descent (SGD) algorithm [28] to train the model and take Tensorflow and Keras as our experiment architectures. The training of the deep network is implemented on TITAN Xp. The batch size of DRN is set to 64. The epoch size is 10. The initial learning rate is set to 0.001, which will be adjusted at each epoch with algorithm of 0.005*0.4**epoch, where "**" is power calculation. The weight decay is set to 0.0001. We add a dropout layer after down-sampling block and set the dropout rate to 0.2 to prevent overfitting. We use the value of test loss as a metric, and the training process stops when the loss values do not descend. The weights are kept when the training of each epoch finishes.

**Character string recognition results.** We show the recognition results of DRN and other compared methods in Table III, where "Accuracy" refers to the correct proportion of whole string and statistics on the test set. For each model, the results are based on the frameworks of CRNN/CNN+CTC. More specifically, the frameworks with the suffix "res-blstm" denotes the models with blstm [29] in a form of residuals, the frameworks with the suffix "no-blstm" means that there is no LSTM layer. "DenseNet-sum-blstm-full-res-blstm" has two changes compared with "Densenet-res-blstm": (1) the approach of combining the two lstms into blstm changes from concat to sum; (2)

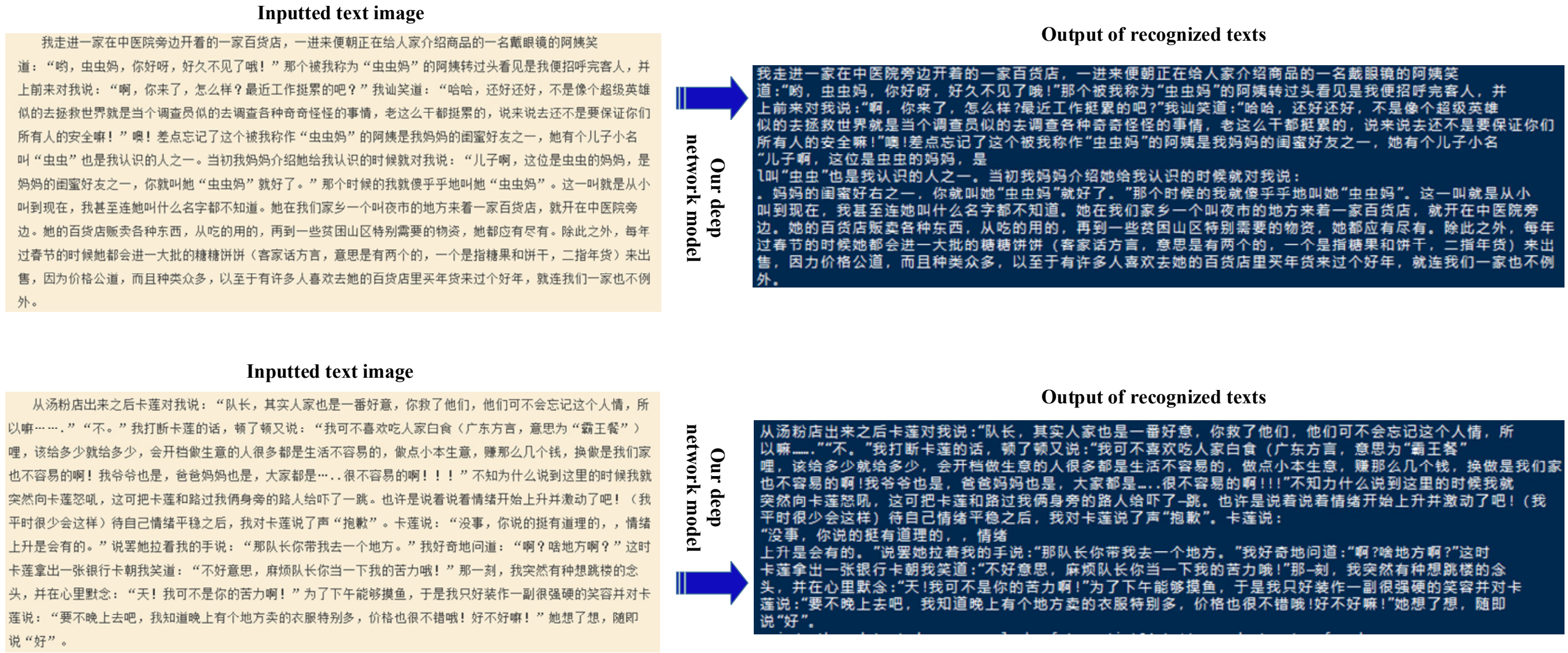

Fig. 8. Illustration of some Chinese recognition results using our proposed DRN framework.

both layers of blstm are connected by residual way. "DenseNet-no-blstm-vertical-feature" removes the pooling operation [30] of 1x4 to "Densenet-no-blstm" relatively. "DenseNet-UB" denotes the DenseNet framework with an up-sampling block, where bilinear interpolation is used to construct the up-sampling block [34]. We see that our proposed DRN delivers higher accuracy and efficiency, compared with the other methods, and the improvement is about 0.42%-7.09%. The results once again show that DRN can deliver better recognition results. The used dataset and most compared methods are publicly available at https://github.com/senlinuc/caffe_ocr.

**Visualization of recognized texts in images.** In addition to the above quantitative results, we also visualize some recognized texts in images using our DRN in Fig.8. To visualize the recognized texts, we employ the Connectionist Text Proposal Network (CTPN) [35] to extract the key text lines from the test images. We can see that our network can output high-quality recognition results. Note that for some identified sentences, there have some deviations in text positions, which is due to the fact that CTPN has no layout analysis function making it fail to produce accurate text alignment when detecting text lines.

## VI. Conclusion and Future Work

In this paper, we investigated the representation learning problem that towards local residual dense learning to global dense residual learning. Technically, we propose a new dense residual network (DRN) for text image representation and recognition. The refined residual dense block (r-RDB) and the global dense block serve as the basic modules of our DRN, where r-RDB can not only retain the advantages of residual dense block , i.e., local feature fusion and residual learning, but also refines the block structures to reduce the computing cost of inner layers. To ensure maximum global information flow between blocks, GDB learns the global dense residual features fully. We also use two convolution layers with stride 2 and more channels to reduce the global feature size and extract more informative deeper features.

We evaluated the performance of DRN for character string and handwritten character recognition. From the investigated cases, although enhanced performance have been obtained by DRN, compared with other related deep models, some issues are still worthy of exploring. For example, more efficient ways to reduce the computing cost and weight size of network, and more effective ways to explore global features are highly-desired to be studied. It is also interesting to use the presented r-RDB and GDB for other popular deep convolutional networks or evaluate DRN for other popular low-level or high-level vision tasks, for instance image restoration and object tracking [31-32]. Besides, how to design an end-to-end text line extraction and recognition framework will also be investigated in our future work.

## Acknowledgments

This work is partially supported by the National Natural Science Foundation of China (61672365, 62072151, 61806035 and U1936217), Anhui Provincial Natural Science Fund for Distinguished Young Scholars (2008085J30) and the Fundamental Research Funds for Central Universities of China (JZ2019HGPA0102). Both Dr. Yang Wang and Dr. Choujun Zhan are the co-corresponding authors of this paper.

## References

[1] F. Zhan and S. Lu, "Esir: End-to-end scene text recognition via iterative image rectification," In: *CVPR*, Long Beach, USA, pp. 2059-2068, 2019.

[2] Z. Cheng, Z. Xu, F. Bai, Y. Niu, S. Pu and S. Zhou, "Aon: Towards arbitrarily-oriented text recognition," In: *CVPR*, Salt Lake City, USA, pp. 5571-5579, 2018.

[3] W. Li, L. Cao, D. Zhao and X. Cui, "CRNN: Integrating classification rules into neural network," In: *Proceedings of the International Joint Conference on Neural Networks,* Dallas, Texas, USA, pp. 1-8, 2013.

[4] G. Huang, Z. Liu, L. Van Der Maaten, and K.Q. Weinberger, "Densely Connected Convolutional Networks," In: *CVPR*, Honolulu, Hawaii, pp.2261-2269, 2017.

[5] K. Cho, B. Van Merriënboer, C. Gulcehre, D. Bahdanau, F. Bougares, H. Schwenk, and Y. Bengio, "Learning phrase representations using RNN encoder-decoder for statistical machine translation," In: *Proceedings of the International Conference on Empirical Methods in Natural Language Processing*, Lisbon, Portugal, pp.1724-1734, 2014.

[6] A. Graves, A.R. Mohamed and G. Hinton, "Speech recognition with deep recurrent neural networks," In: *ICASSP*, Vancouver, British Columbia, Canada, pp. 6645-6649, 2013.

[7] C. McBride-Chang, H. Shu, A. Zhou, C. Wat and R. Wagner, "Morphological awareness uniquely predicts young children's Chinese character recognition," *Journal of Educational Psychology*, vol.95, No.4, pp.743, American Psychological Association, 2003.

[8] A. Graves, S. Fernández, F. Gomez and J. Schmidhuber, "Connectionist temporal classification: labelling unsegmented sequence data with recurrent neural networks," In: *ICML*, Pittsburgh, Pennsylvania, USA, pp.369-376, 2006.

[9] Y. Gao, Y. Chen, J. Wang and H. Lu, "Reading scene text with attention convolutional sequence modeling," *Neurocomputing*, vol.339, pp.161-170, 2017.

[10] K. He, X. Zhang, S. Ren and J. Sun, "Deep residual learning for image recognition," In: *CVPR*, Las Vegas, USA, pp. 770-778, 2016.

[11] Y. Zhang, Y. Tian, Y. Kong, B. Zhong, Y. Fu, "Residual dense network for image super-resolution," In: *CVPR*, Utah, USA, 2018.

[12] Z. Zhang, Z.M. Tang, Z. Zhang, Y. Wang, J. Qin and M. Wang, "Fully-Convolutional Intensive Feature Flow Neural Network for Text Recognition," In: *ECAI*, Santiago de Compostela, June 2020.

[13] S. Ahmed, M. I. Razzak and R. Yusof, "Cursive Script Text Recognition in Natural Scene Images - Arabic Text Complexities," *Springer*, ISBN 978-981-15-1296-4, pp.1-111, 2020.

[14] W. He, X. Zhang, F. Yin, Z. Luo, J. Ogier, C. Liu, "Realtime multi-scale scene text detection with scale-based region proposal network, *Pattern Recognition*, vol.98, pp.107026, 2020.

[15] H. Caulfield and W. Maloney, "Improved discrimination in optical character recognition," *Applied Optics*, vol.8, No.11, pp.2354-2356, 1969.

[16] Z. Zhang, Z. Tang, Y. Wang, H. Zhang, S. Yan, M. Wang, "Compressed DenseNet for Lightweight Character Recognition," *arXiv:1912.07016 [cs.CV]*, 2020.

[17] S. Ioffe, C. Szegedy, "Batch normalization: Accelerating deep network training by reducing internal covariate shift," In: *ICML*, 2015.

[18] K. Hara, D. Saito and H. Shouno, "Analysis of function of rectified linear unit used in deep learning," In: *IJCNN*, Killarney, Ireland, pp. 1-8, 2015.

[19] A.G. Howard, M. Zhu, B. Chen, D. Kalenichenko, W. Wang, T.Weyand, and H. Adam, "Mobilenets: Efficient convolutional neural networks for mobile vision applications," *arXiv preprint arXiv:1704.04861*, 2017.

[20] K. Duan, S. Keerthi, W. Chu, S. Shevade, and A. Poo, "Multi-category Classification by Soft-Max Combination of Binary Classifiers," In: *Proceedings of the Multiple Classifier Systems, International Workshop*, Guilford, Uk, 2003.

[21] Y. Jia, E. Shelhamer, J. Donahue, S. Karayev, J. Long, R. Girshick, and T. Darrell, "Caffe: Convolutional architecture for fast feature embedding," In: *ACM Multimedia*, Orlando, FL, USA, pp. 675-678, 2014.

[22] A. Gupta, A. Vedaldi, A. Zisserman, "Synthetic data for text localisation in natural images," In: *CVPR*, Las Vegas, USA, pp. 2315-2324, 2016.

[23] Y. LeCun, C. Cortes and C. Burges, "MNIST handwritten digit database," AT&T Labs, Available: http://yann.lecun. com/exdb/mnist, 2010.

[24] Y. Tang, "Deep learning using linear support vector machines," In: *ICML Workshops*, Atlanta, Georgia, USA, 2013.

[25] I. Goodfellow, D. Warde-Farley, M. Mirza, A. Courville, and Y. Bengio, "Maxout networks," In: *ICML,* Atlanta, GA, USA, 2013.

[26] M. Courbariaux, Y. Bengio and J. David, "Binaryconnect: Training deep neural networks with binary weights during propagations," In: *NIPS*, pp. 3123-3131, 2015.

[27] T.H. Chan, K. Jia, S. Gao, J. Lu, Z. Zeng, and Y. Ma, "PCANet: A simple deep learning baseline for image classification?" *IEEE Transactions on Image Processing*, vol. 24, No. 12, pp. 5017-5032, 2015.

[28] Z.H. Zhou, and J. Feng, "Deep Forest: Towards An Alternative to Deep Neural Networks," In: *IJCAI*, Melbourne, Australia, pp.3553-3559, 2017.

[29] G. Lefebvre, S. Berlemont, F. Mamalet, and C. Garcia, "BLSTM-RNN based 3D gesture classification," In: *ICANN*, Berlin, pp.381-388, 2013.

[30] G. Stasser, W. Titus, "Pooling of unshared information in group decision making: Biased information sampling during discussion," *Journal of personality and social psychology*, vol. 48, No. 6, pp. 1467, 1985.

[31] W. Dong, P. Wang, W. Yin, G. Shi, F. Wu, and X. Lu, "Denoising Prior Driven Deep Neural Network for Image Restoration," *IEEE Trans. on Pattern Anal. Mach. Intell.,* vol.41, no.10, pp.2305-2318, 2019.

[32] B. N. Vo, B. Vo and M. Beard, "Multi-sensor multi-object tracking with the generalized labeled multi-Bernoulli filter," *IEEE Transactions on Signal Processing*, vol. 67, No. 23, pp. 5952-5967, 2019.

[33] J. Wang and X. Hu, "Gated Recurrent Convolution Neural Network for OCR," In: *NIPS*, pp.35-344, 2017.

[34] Z. Tang, W. Jiang, Z. Zhang, M. Zhao, L. Zhang, M. Wang, "DenseNet with Up-Sampling Block for Recognizing Texts in Images," *Neural Computing and Applications*, vol.32, no.11, pp.7553-7561, 2020.

[35] Z. Tian, W. Huang, T. He, P. He, and Y. Qiao, "Detecting text in natural image with connectionist text proposal network," In: *ECCV*, Amsterdam, The Netherlands, pp. 56-72, 2016.

[36] X. Bai, C. Yao, W. Liu, "Strokelets: A Learned Multi-Scale Mid-Level Representation for Scene Text Recognition," *IEEE Trans. on Image Processing*, vol.25, no.6, pp.2789-2802, 2016.

[37] M. Liao, J. Zhang, Z. Wan, F. Xie, J. Liang, P. Lyu, C. Yao and X. Bai, "Scene Text Recognition from Two-Dimensional Perspective," In: *AAAI Conference on Artificial Intelligence*, pp.8714-8721, 2019.

[38] Y. Zhang, S. Nie, W. Liu, X. Xu, D. Zhang and H. Shen. "Sequence-To-Sequence Domain Adaptation Network for Robust Text Image Recognition," In: *CVPR*, Long Beach, CA, pp.2740-2749, 2019.

[39] N. Acuff and J. Linden, "Using visualization of t-distributed stochastic neighbor embedding to identify immune cell subsets in mouse tumors," *Journal of Immunology*, vol.198, No.11, pp.4539-4546, 2017.

[40] L. Breiman, "Random forests," *Machine learning*, vol. 45, No. 1, pp. 5-32, 2001.

[41] H. Bourlard and C. J. Wellekens, "Links between Markov models and multilayer perceptrons," *IEEE Transactions on Pattern Analysis and Machine Intelligence*, vol. 12, No. 12, pp. 1167-1178, 1990.

[42] F. Pedregosa, G. Varoquaux, A. Gramfort, V. Michel, B. Thirion, O. Grisel, M. Blondel, P. Prettenhofer, R. Weiss, V. Dubourg, et al., "Scikit-learn: Machine learning in python," *Journal of Machine Learning Research*, vol. 12, pp. 2825–2830, 2011.

[43] M. Thoma, "The hasyv2 dataset," *arXiv preprint arXiv:1701.08380*, 2017.

[44] Y. Zheng, B. K. IWANA, and S. UCHIDA, "Mining the displacement of max-pooling for text recognition," *Pattern Recognition*, vol. 93, pp. 558-569, 2019.

[45] Z. Zhang, Y. Sun, Z. Zhang, Y. Wang, L. Wu and M. Wang, "MDPL-net: Multi-layer Dictionary Learning Network with Added Skip Dense Connections," In: *ICDM*, Sorrento, Italy, August 2020.

[46] Y. Jia, H. Zhang, Z. Zhang and M. Liu, "CNN-based Encoder-Decoder Networks for Salient Object Detection: A Comprehensive Review and Recent Advances," *Information Sciences*, vol.546, pp.835-857, Feb 2021.

[47] G. S. Xie, Z. Zhang, L. Liu, F. Zhu, X.Y. Zhang, L. Shao, X. Li, "SRSC: Selective, Robust, and Supervised Constrained Feature Representation for Image Classification," *IEEE Trans. Neural Networks Learn. Syst.*, vol.31, no.10, pp.4290-4302, 2020.

[48] G. Xie, X. Zhang, S. Yan, C. Liu, "Hybrid CNN and Dictionary-Based Models for Scene Recognition and Domain Adaptation," *IEEE Trans. Circuits Syst. Video Technol.*, vol.27, no.6, pp.1263-1274, 2017.

[49] X. Gao, Z. Zhang, T. Mu, X. Zhang, C. Cui and M. Wang, "Self-attention Driven Adversarial Similarity Learning Network," *Pattern Recognition*, vol.105, pp.107331, Sep 2020.

[50] P. P. Angelov and E. Soares, "Towards explainable deep neural networks (xDNN)," *Neural Networks*, vol.130, pp.185-194, 2020.

[51] D. Bacciu, F. Errica, A. Micheli and M. Podda, "A gentle introduction to deep learning for graphs," *Neural Networks*, vol.129, pp.203-221, 2020.

[52] H. Mhaskar and T. A. Poggio, "An analysis of training and generalization errors in shallow and deep networks," *Neural Networks*, vol.121, pp.229-241, 2020.